\documentclass{article}

\usepackage{arxiv}

\usepackage[utf8]{inputenc} 
\usepackage[T1]{fontenc}    
\usepackage{hyperref}       
\usepackage{url}            
\usepackage{booktabs}       
\usepackage{amsfonts}       
\usepackage{nicefrac}       
\usepackage{microtype}      
\usepackage{lipsum}
\usepackage{graphicx}
\usepackage{xargs}
\usepackage{textgreek}
\usepackage{hyperref}

\title{Neuromodulated Patience for Robot and Self-Driving Vehicle Navigation}

\author{
  Jinwei Xing \\
  Department of Cognitive Sciences\\
  University of California, Irvine\\
  Irvine, CA 92697-5100 \\
  \texttt{jinweix1@uci.edu} \\
   \And
  Xinyun Zou \\
  Department of Computer Science\\
  University of California, Irvine\\
  Irvine, CA 92697-5100 \\
  \texttt{xinyunz5@uci.edu} \\
  \And
  Jeffrey L. Krichmar\\
  Department of Cognitive Sciences\\
  Department of Computer Science\\
  University of California, Irvine\\
  Irvine, CA 92697-5100 \\
  \texttt{jkrichma@uci.edu} \\
}

\begin{document}
\maketitle

\begin{abstract}
Robots and self-driving vehicles face a number of challenges when navigating through real environments. Successful navigation in dynamic environments requires prioritizing subtasks and monitoring resources. Animals are under similar constraints. It has been shown that the neuromodulator serotonin regulates impulsiveness and patience in animals. In the present paper, we take inspiration from the serotonergic system and apply it to the task of robot navigation. In a set of outdoor experiments, we show how changing the level of patience can affect the amount of time the robot will spend searching for a desired location. To navigate GPS compromised environments, we introduce a deep reinforcement learning paradigm in which the robot learns to follow sidewalks. This may further regulate a tradeoff between a smooth long route and a rough shorter route. Using patience as a parameter may be beneficial for autonomous systems under time pressure.
\end{abstract}


\section{Introduction}
Real-world environments can change due to the season, time of day, construction, or the behavior of other agents. Uncertainty can arise due to sensor noise, unforeseen obstacles or uncertain goals.  An autonomous system needs to cope with these challenges and have the ability to adapt its behavior based on the current situation.

Successful behavior requires a tradeoff between patience and assertiveness. For example, a self-driving car may get stuck at a four-way stop sign because human drivers are not waiting their turn. Or an autonomous robot whose task is to thoroughly explore an environment may follow this directive even if it runs out of power. In both cases, a signal regulating the patience of the system would be beneficial. The self-driving car would eventually assert itself, and the navigating robot would quit its exploration and seek a charging station.

The brain has a number of neuromodulators that regulate context, signal changes, and direct actions. The neuromodulator serotonin (5-HT) is thought to have a role in harm aversion, anxious states, and temporal discounting \cite{RN18,rn19,RN20}. One theory posits that serotonin regulates the impulsiveness of action. Recently, Miyazaki and colleagues showed that optogenetically increasing serotonin levels caused mice to be more patient, especially when the timing of a reward was uncertain \cite{RN13}. Based on these results, they developed Bayesian decision model for the probability to wait or quit.

In the present paper, we apply this model of impulsiveness to robot navigation. Specifically, the robot will navigate through a series of waypoints. The level of serotonin dictates how patiently the robot will search for a waypoint. We show that changing the serotonin level can have dramatic effects on the robot's behavior.  Such a system may be beneficial for adjusting autonomous behavior depending on the dynamics and uncertainty of the environment.  

\section{Methods}
\label{sec:methods}

\subsection{Navigation task}
\begin{figure}
  \centering
  \includegraphics[width=1\linewidth]{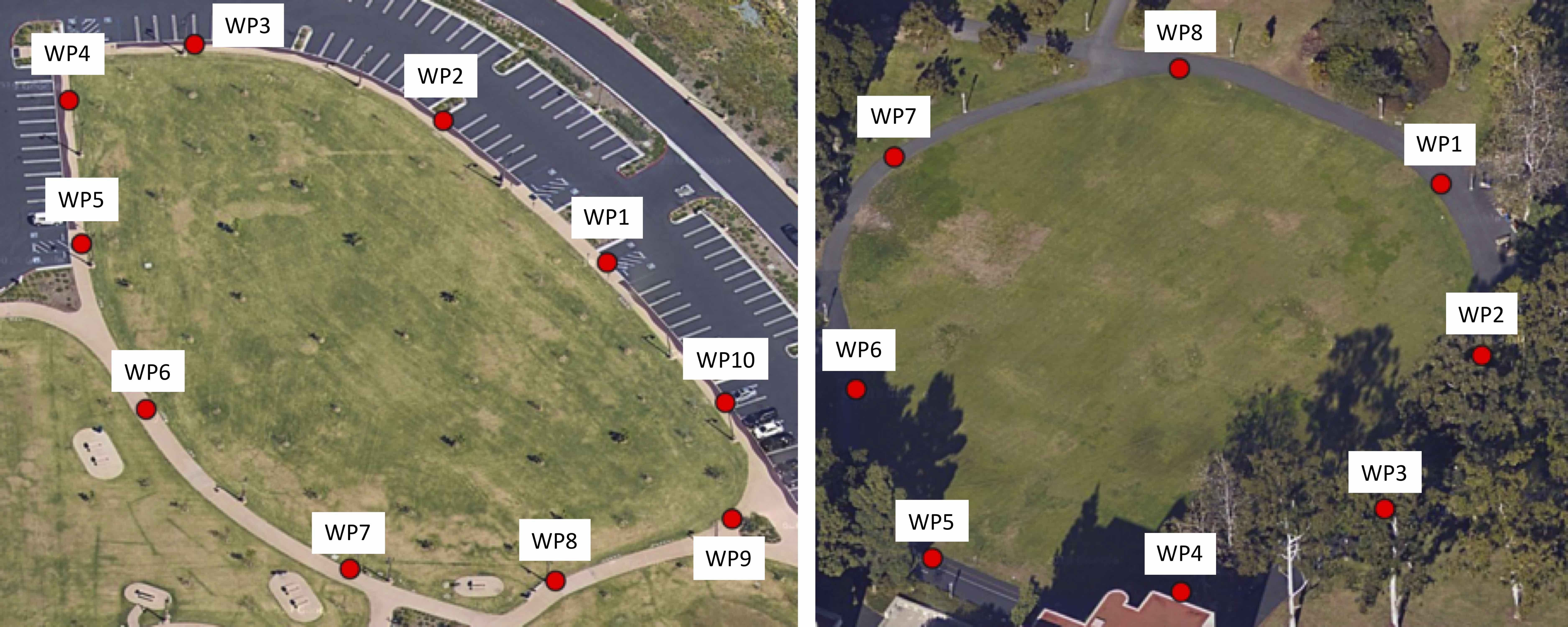}
  \caption{Parks at which robot navigation experiments were carried out. The left is an image of Encinitas Community park and the right is an image of Aldrich park at the University of California, Irvine. The labels denote the waypoints. Waypoints were approximately 50-60 meters apart. Imagery from Google Maps, 2019.}
  \label{fig:figParks}
\end{figure}

A set of robot navigation tasks were carried out in two different parks. Figure \ref{fig:figParks} shows satellite images of the two parks. The waypoints for each park are superimposed on the maps. Waypoints were GPS coordinates placed on sidewalks in the park. The park on the left of Figure \ref{fig:figParks}, Encinitas Community park, was relatively flat. The park on the right of Figure \ref{fig:figParks}, Aldrich park at the University of California, Irvine,  was hilly with numerous obstacles. It also should be noted that Aldrich park was in a sunken bowl surrounded by tall university buildings. These features made GPS signals unreliable. 

In both parks, the robot was to proceed to each waypoint in order. If the robot became impatient, it would skip searching for the present waypoint and randomly choose a future waypoint. However, the robot had to reach the last waypoint for a trial to be complete.

\subsection{Robot and Software Design}

\begin{figure}
  \centering
  \includegraphics[width=1\linewidth]{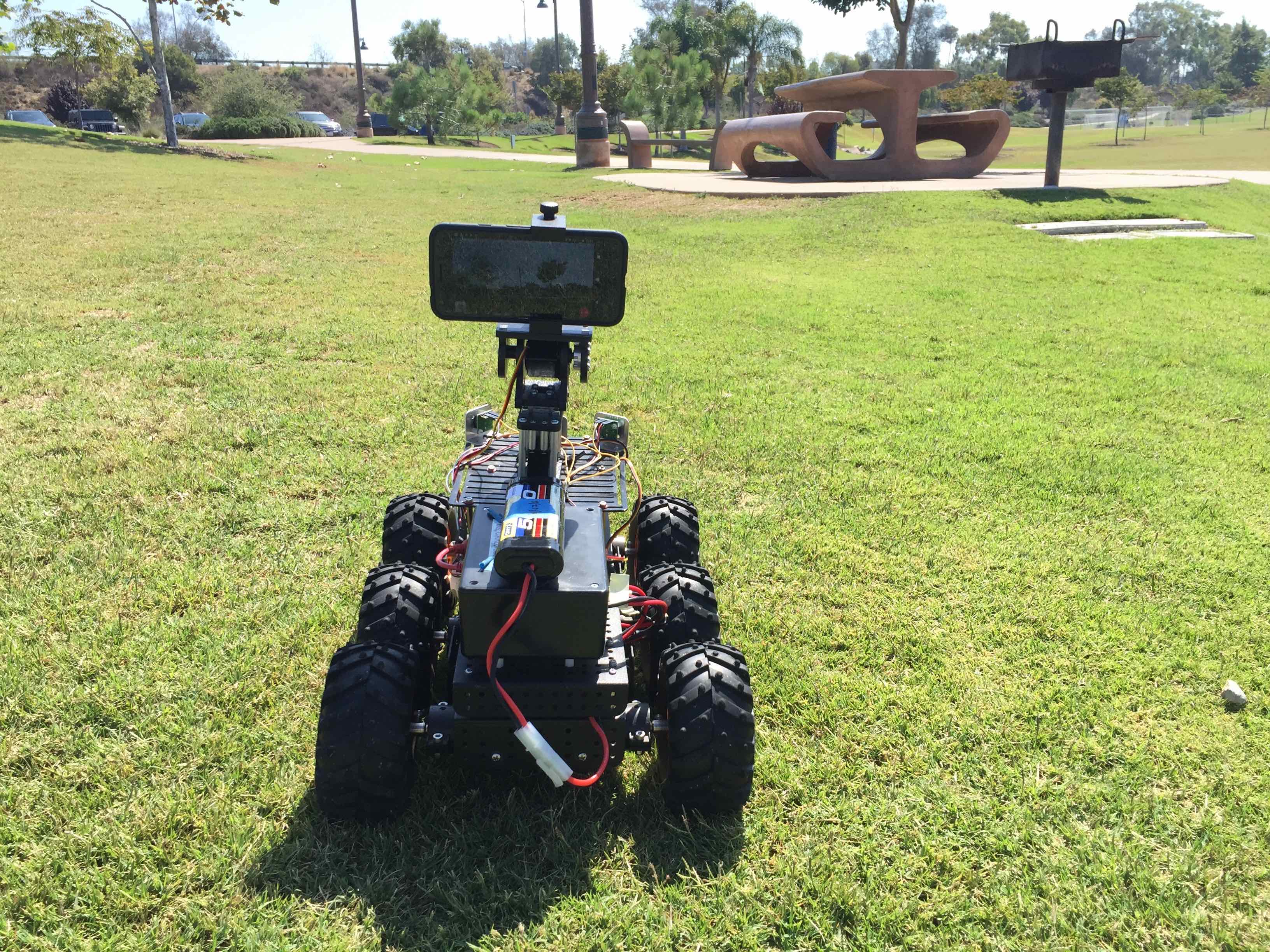}
  \caption{Android based robot used for the experiments.}
  \label{fig:figRobot}
\end{figure}

For the robot experiments, we used the Android-based robotic platform \cite{RN15,rn16,RN17}, a mobile ground robot constructed from off-the-shelf commodity parts and controlled through an Android smartphone (see Figure \ref{fig:figRobot}). An IOIO-OTG microcontroller communicated with an Android smartphone via a Bluetooth connection and relayed motor commands to a separate motor controller for steering the Dagu Wild Thumper 6-Wheel Drive All-Terrain chassis. Three ultrasonic sensors, which were used for obstacle avoidance, were connected to the robot through the IOIO-OTG. The software application was written in Java using Android Studio and deployed on a Google Pixel XL smartphone. The application utilized the phone’s built-in camera, accelerometer, gyroscope, compass, and GPS for navigation.

For waypoint navigation, a GPS location was queried using the Google Play services location API. The bearing direction from the current GPS location of the robot to a desired waypoint was calculated using the Android API function bearingTo. A second value, the heading, was calculated by subtracting declination of the robot’s location to the smartphone compass value, which was relative to magnetic north. This resulted in an azimuth direction relative to true north. The robot traveled forward and steered in attempt to minimize the difference between the bearing and heading. The steering direction was determined by deciding whether turning left or turning right would require the least amount of steering to match the bearing and heading. The navigation procedure continued until the distance between the robot’s location and the current waypoint was less than 20 meters, at which point the next waypoint in the list was selected.

\subsection{Waypoint Navigation and Model of Impulsivity}

The robot proceeded through a list of waypoints as described above. However, if the robot became impatient, it skipped the present waypoint and randomly chose a waypoint closer to the final destination. 

\begin{figure}
  \centering
  \includegraphics[width=1\linewidth]{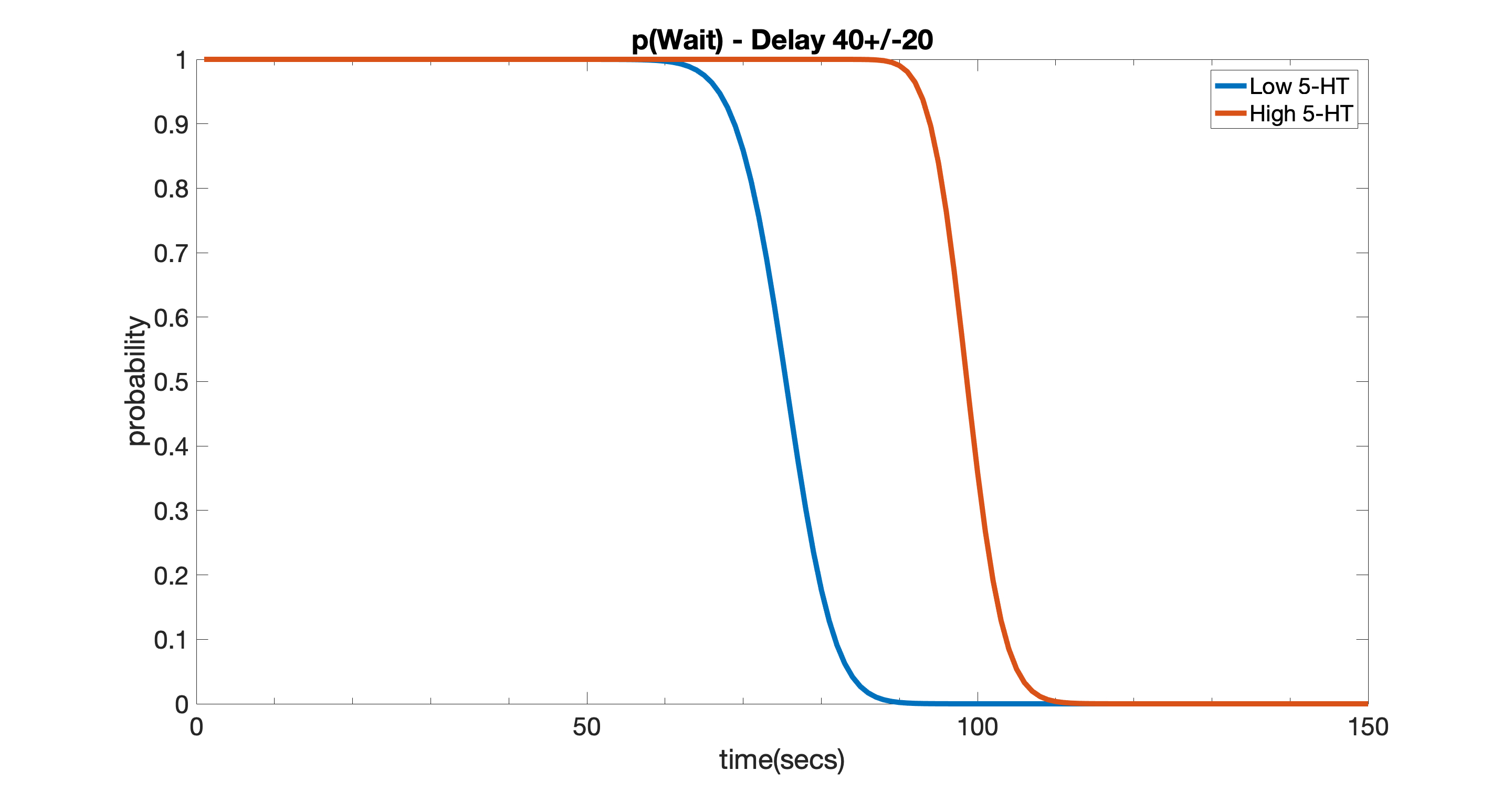}
  \caption{Probability of waiting function. Higher serotonin (5-HT) levels shifted the curve to the right resulting in longer wait times.}
  \label{fig:figWait}
\end{figure}

The likelihood to skip a waypoint was based on the Bayesian Decision Model given by \cite{RN13}. Specifically, we calculated the probability to wait given the time elapsed.

\begin{equation}
p(wait|t)= {\frac {1}{1+exp^{\beta 5HT(t)}}}
\end{equation}

where {\textbeta}  was equal to 50 and 5HT(t) was the likelihood of reaching the waypoint at time $t$. The likelihood was calculated with a Normal cumulative distribution function having a mean of 40 seconds and a standard deviation of 20 seconds. The likelihood function was multiplied by a scalar that represented the probability of receiving a reward. As in \cite{RN13}, we assume that increasing serotonin levels causes an overestimation of the prior probability. Therefore, in our experiments low serotonin equated to a probability of a reward of 0.50 and high serotonin equated to probability of a reward of 0.95 (see \cite{RN13} for details). Figure \ref{fig:figWait} shows the resulting probability to wait, $p(Wait|t)$, curves.

The $p(Wait|t)$ curves in Figure \ref{fig:figWait} were used to decide whether to keep searching for a waypoint, or forego the desired waypoint and choose another. A random number between 0 and 1 was generated and if the number was greater than $p(Wait|t)$, where $t$ is the time elapsed searching for a waypoint, the robot stopped searching for this waypoint.  A new waypoint was randomly chosen that was closer to the final destination. Note that if the robot was searching for the final destination waypoint or for a new waypoint after a skip, the $p(Wait|t)$ curve was not referenced. That is, the robot had to reach the shortcut waypoint and had to reach the final waypoint for a successful trial.

\subsection{Road Following with Deep Reinforcement Learning}

\begin{figure}
  \centering
  \includegraphics[width=1\linewidth]{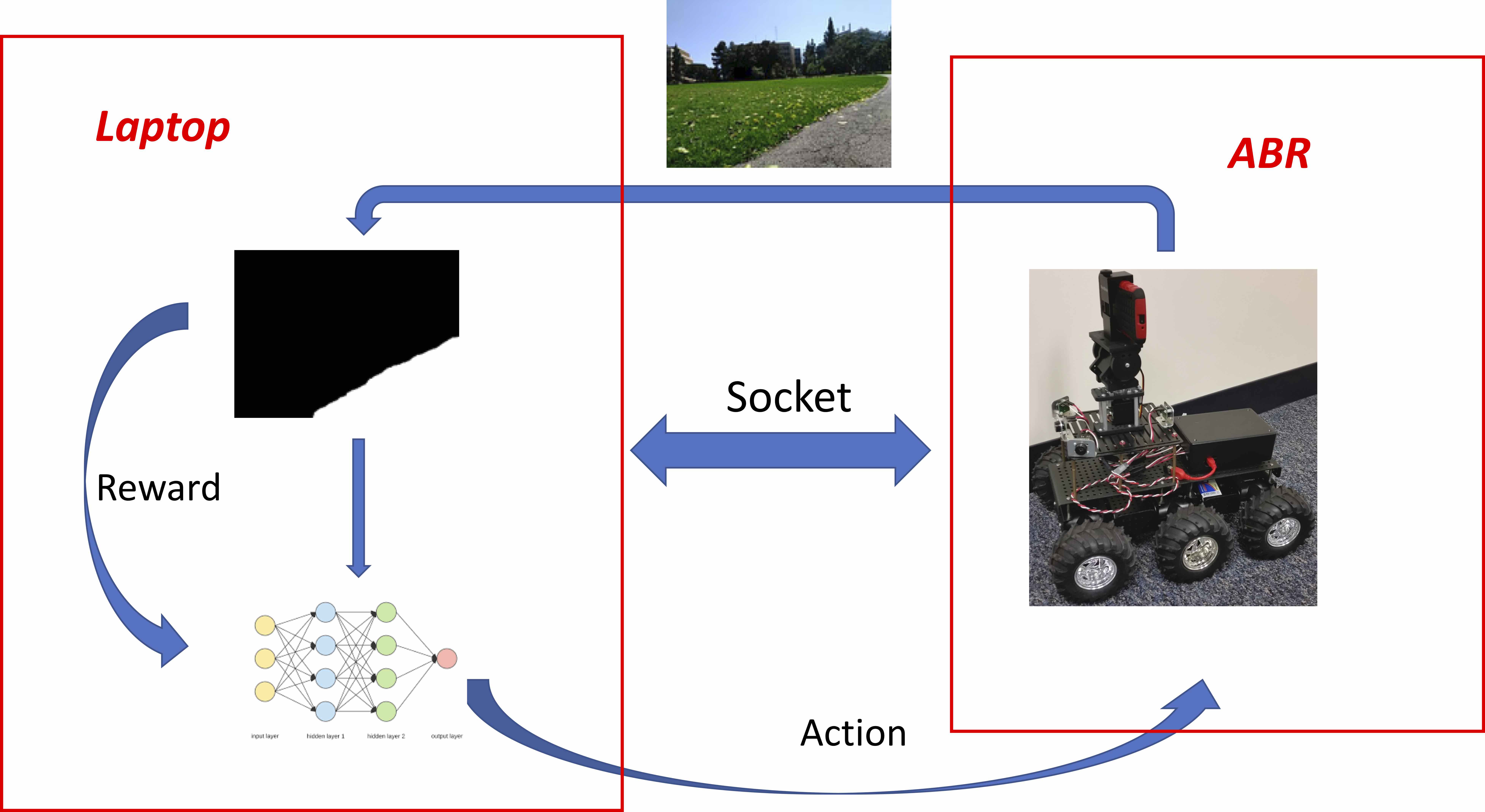}
  \caption{High level illustration of the data pipeline for the road following algorithm. Images from the Android based robot's smartphone camera are sent to a nearby laptop via a socket. The laptop runs a deep reinforcement learning algorithm, which rewards staying on the road, and generates steering actions for the robot. }
  \label{fig:dataPipe}
\end{figure}

A road following algorithm was used in some of the experiments carried out in Aldrich park. We used deep reinforcement learning models for online learning of a driving policy on the Aldrich park sidewalks. Figure \ref{fig:dataPipe} shows the data pipeline. The Android Based Robot took pictures with the smartphone’s camera. Using a “hotspot”, the image was sent to a nearby laptop, which performed image segmentation of “road” vs. “non-road”. The laptop had a deep reinforcement learning network, based on DQN \cite{RN21}, which processed the image, and suggested an action. The actions ranged from sharp left to slight left to straight to slight right and to sharp right. The robot got a reward if it stayed on the road and a penalty if it went off the road. The laptop took about 400 ms to process the information, generate an action, and update the network. This was adequate for online learning in real-time. 

To provide road information to the network, we used a semantic segmentation neural network \cite{RN22}. This allowed us to rapidly label road and non-road portions of a scene. The network was then  trained to generalize this information for scenes the robot viewed in Aldrich park. The robot learned to follow the road after roughly 500 training trials. For now, we are segmenting road and non-road. But, potentially, we could segment people, trees, benches, etc. These object classes could be used as further inputs for training the network. 

\begin{figure}
  \centering
  \includegraphics[width=1\linewidth]{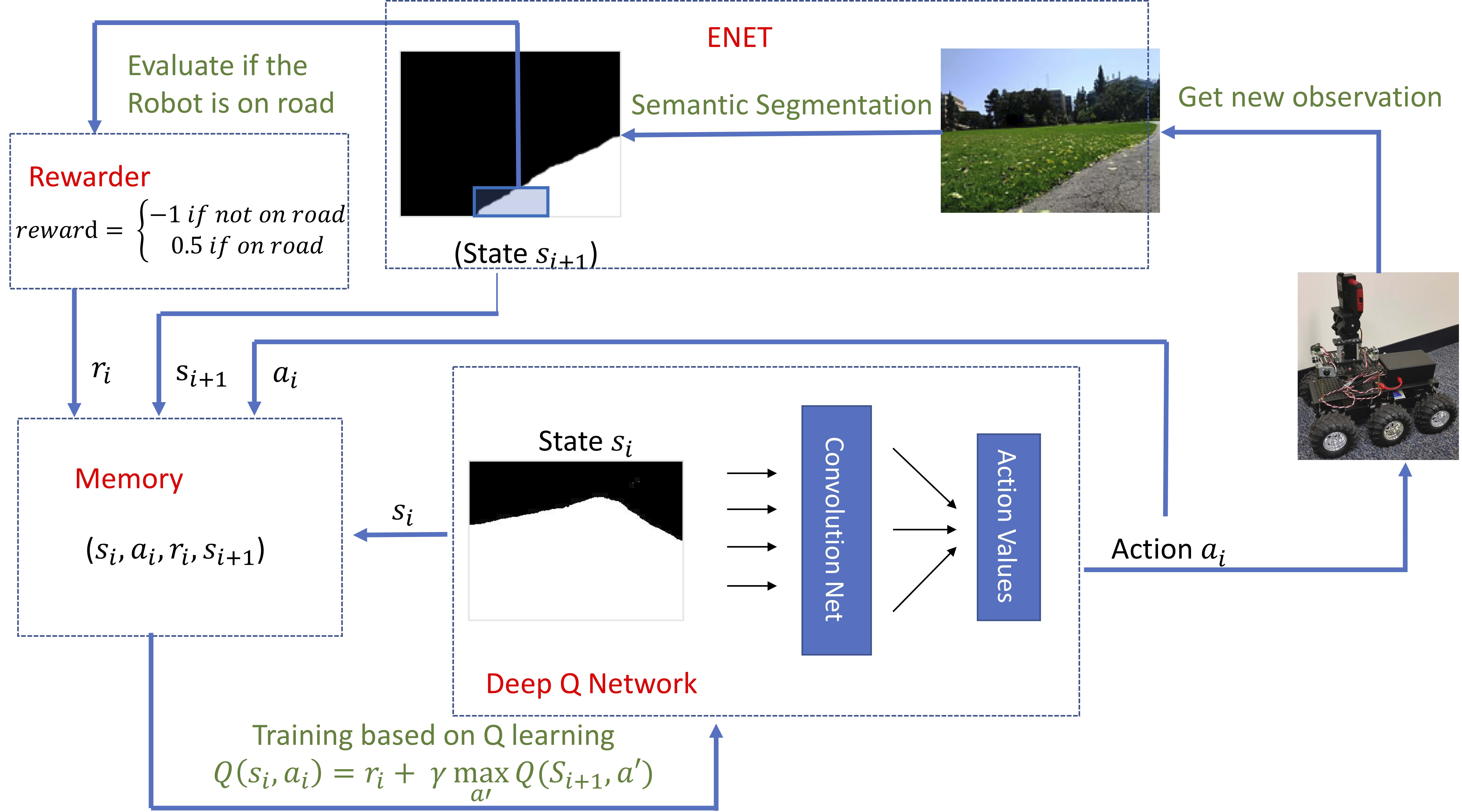}
  \caption{Detailed illustration of the data pipeline for the road following algorithm. Enet was used to segment road from non-road \cite{RN22}. The network gave a positive reward for actions that kept the robot on the road and a penalty for actions that caused the robot to go off road. Training was based on a DQN reinforcement learning paradigm \cite{RN21}. Training and testing were carried out online in Aldrich park.}
  \label{fig:roadRL}
\end{figure}

A detailed illustration of the road following deep reinforcement neural network is given in Figure \ref{fig:roadRL}.

\section{Results}

Two sets of robot navigation experiments were carried. One set was in the Encinitas Community park (see Figure \ref{fig:figParks} left) and the other was in Aldrich park (see Figure \ref{fig:figParks} right). In both cases, the robot navigated through a set of waypoints with low and high serotonin levels. In Aldrich park, an additional navigation experiment was carried out with road following.

\subsection{Waypoint Navigation in Encinitas Community park}

\begin{figure}
  \centering
  \includegraphics[width=1\linewidth]{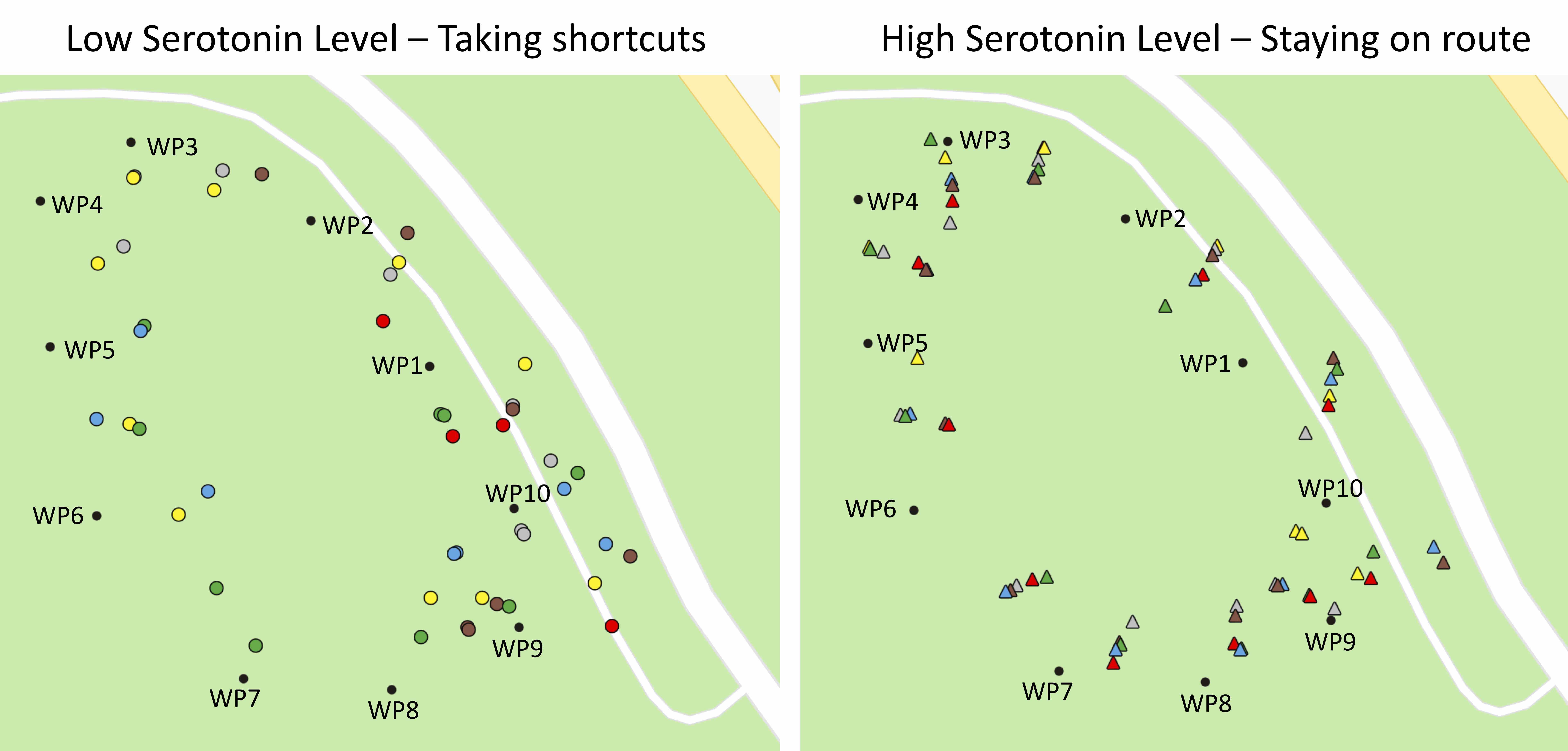}
  \caption{Robot navigation trials in the Encinitas Community park. The black dots are the waypoint destinations (WP1 – WP10). There were 6 trials. Each color represents an individual trial. Each colored marker denotes the robot reaching a waypoint.}
  \label{fig:enc6}
\end{figure}

We ran 6 trials for low serotonin and 6 trials for high serotonin in the Encinitas Community park (see Figure \ref{fig:enc6}). The waypoints were roughly 50-60 meters apart. In Figure \ref{fig:enc6}, each colored marker denotes a trial. The markers denote the GPS location from the smartphone when the robot was within 20 meters of a waypoint. Note that this reading could vary dramatically due to GPS inaccuracies.

The level of serotonin affected the robot's patience in finding a waypoint. Over the 6 trials, 9 waypoints were skipped when serotonin was low, and 2 waypoints were skipped when serotonin was high. The average time before skipping a waypoint was 68 seconds for low serotonin and 97 seconds for high serotonin.  These experiments demonstrated how this model could change route planning behaviors.

Figure \ref{fig:enc2} shows the GPS readings from two trials, one with high serotonin and the other with low serotonin. In the high serotonin trial, the robot reached every waypoint. In the low serotonin trial, the probability to wait was exceeded for reaching waypoint 6 after 69 seconds and the robot skipped to waypoint 9. A video of the robot performing waypoint navigation with low serotonin can be found at: \url{https://youtu.be/6EcNchTGLKw} and a video of the robot performing waypoint navigation with high serotonin can be found at: \url{https://youtu.be/q_m0gbVN6UE}

\begin{figure}
  \centering
  \includegraphics[width=1\linewidth]{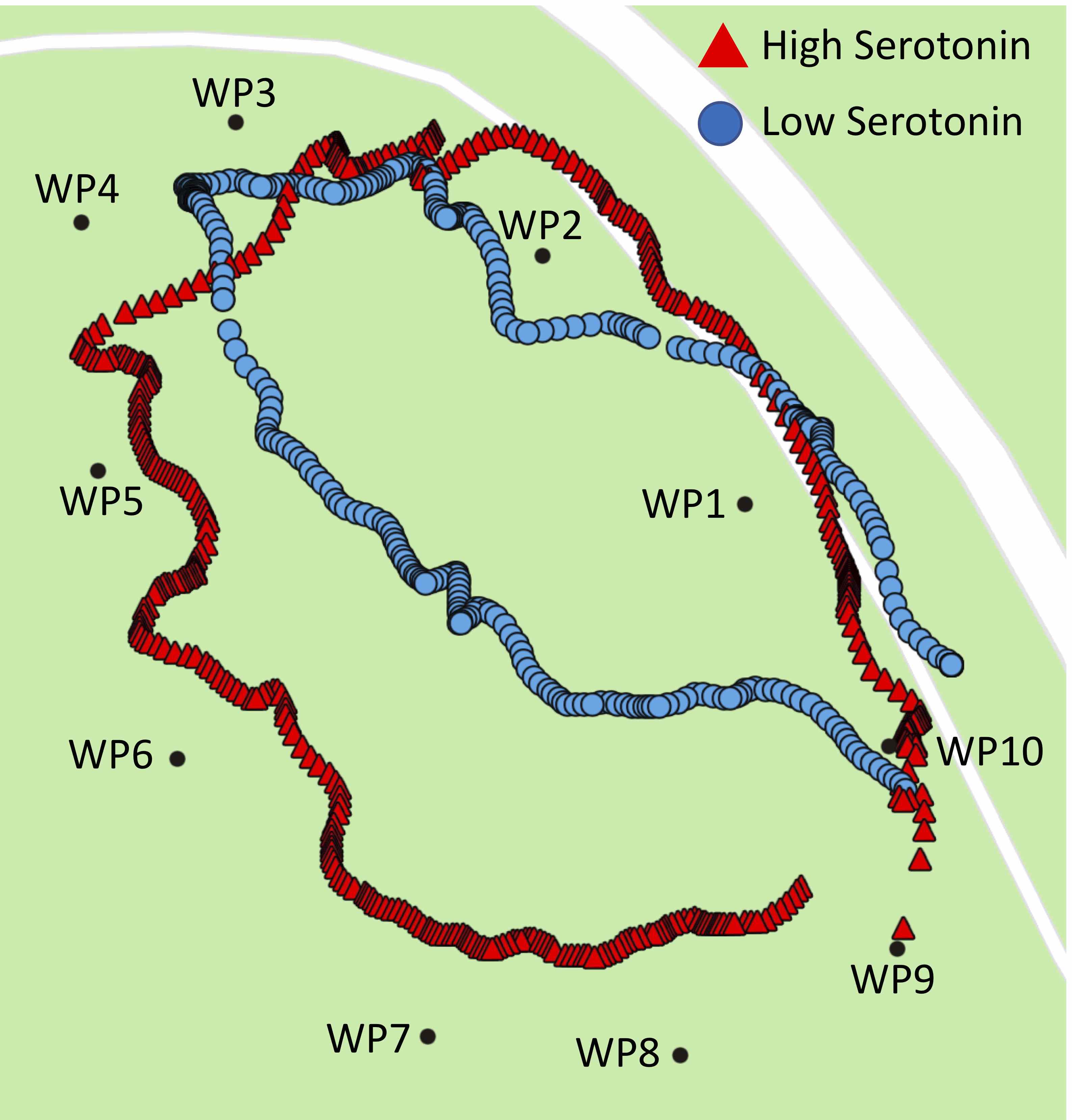}
  \caption{Two representative navigation trials in the Encinitas Community park. All the GPS points are shown. The markers in the figure correspond to the grey markers in Figure \ref{fig:enc6}.}
  \label{fig:enc2}
\end{figure} 

\subsection{Waypoint Navigation in Aldrich park}

\begin{figure}
  \centering
  \includegraphics[width=.5\linewidth]{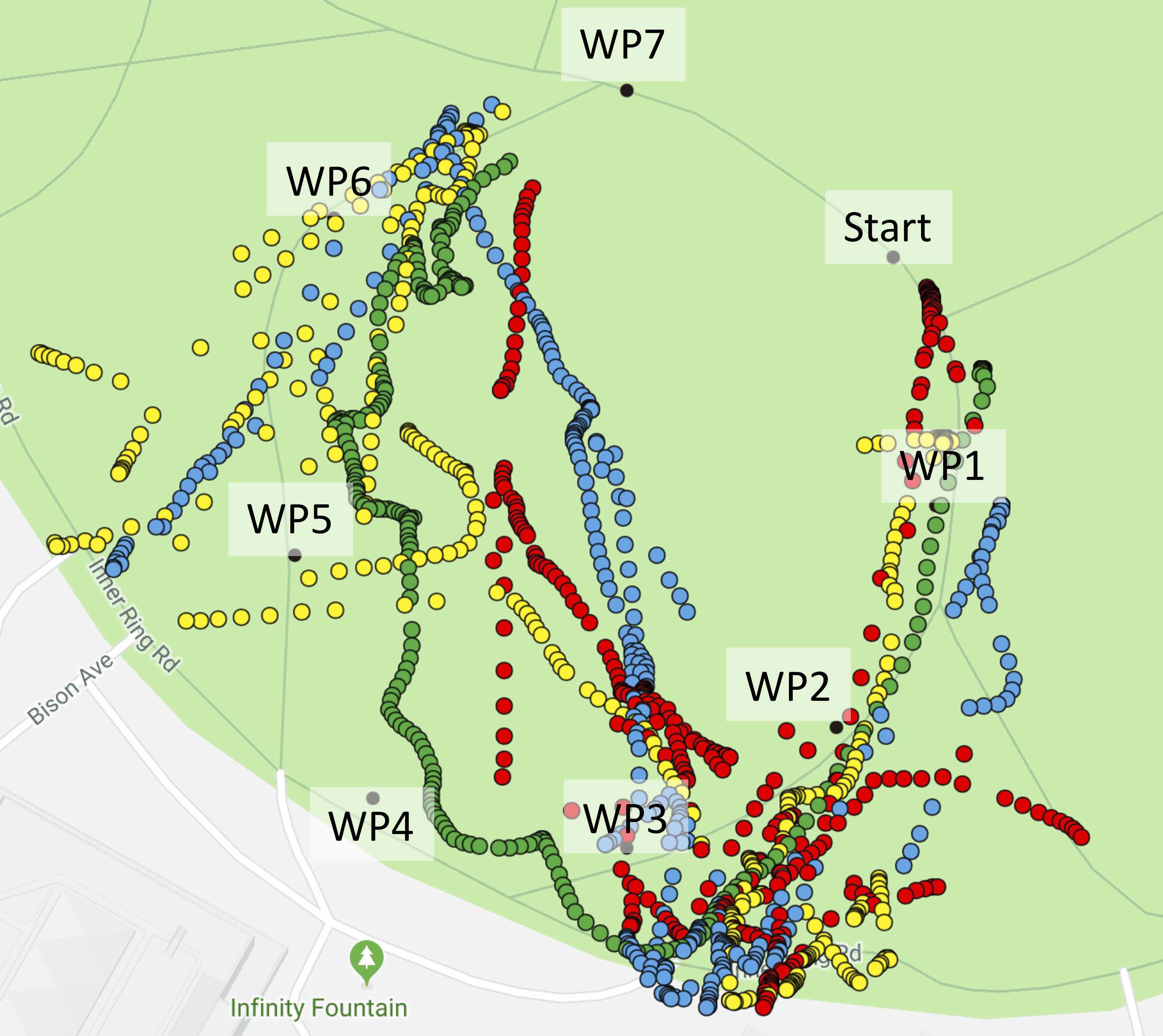}
  \caption{Four trials in Aldrich park with high serotonin.  GPS readings are shown with a colored marker. Different colors denote different trials.}
  \label{fig:aldrich}
\end{figure}

We ran four high serotonin trials in Aldrich park on the University of California, Irvine campus (see Figure \ref{fig:aldrich}). Since the area is sunken in a bowl surrounded by tall buildings and trees, the GPS readings can be highly inaccurate. In particular, the robot had difficulty finding waypoints 2 and 3. On all four trials, after around 98 seconds trying to find a waypoint, the robot took shortcuts to a later waypoint.  

In an additional trial, we introduced the road following algorithm (see Figure \ref{fig:abrroad}). In this trial, the robot stayed on the sidewalk and was able to reach every waypoint within the probability to wait constraint. It should be noted that during road following, the robot took longer to complete the course (6:40) than when shortcuts were taken (6:01 average). However, the robot reached more waypoints and took less energy, since it was traveling over smoother terrain, than the trials when it took shortcuts. These results show the benefit of adding additional navigation tools and the tradeoffs associated with being patient versus being impulsive. A video of the robot navigating using road following can be found at: \url{https://youtu.be/DixOxO2UafQ} 

\begin{figure}
  \centering
  \includegraphics[width=1\linewidth]{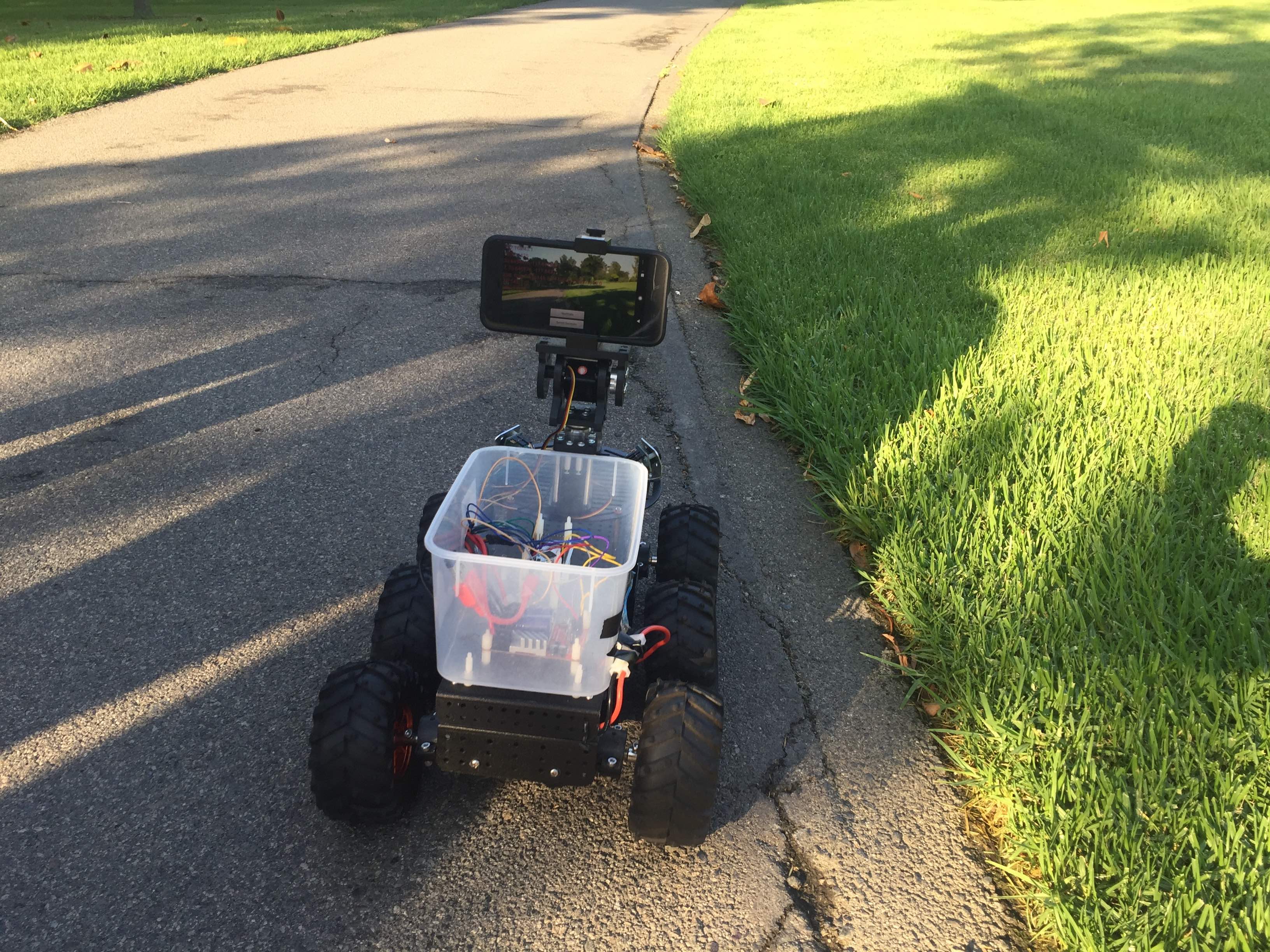}
  \caption{Android based robot using the road following algorithm.}
  \label{fig:abrroad}
\end{figure} 

\section{Discussion}

In the present paper, we showed how a concept from behavioral neuroscience could be applied to robot navigation and possibly self-driving vehicles. It has been shown that serotonin in the brain affects impulsiveness in an animal's behavior \cite{RN13,RN23}. The present model applied this idea to waypoint navigation in autonomous robots. Specifically, we showed that simulating high serotonin led to increased search time for a desired location and that simulating low serotonin led to an increase in calling off the search for some waypoints. Even under high serotonin conditions, if a waypoint was particularly difficult to find or there were environmental challenges, there is a limit to how long the robot will try to reach a desired location (see Figure \ref{fig:aldrich}).

The goal of the present algorithm and demonstrations was not to achieve some benchmark, but rather to suggest a neurobiologically inspired strategy that could complement other navigation systems. The present approach could be applied to biomimetic navigation systems \cite{RN1,RN9}, as well as engineering approaches to navigation \cite{RN2,RN3}. In general, the probability to wait suggests a level of urgency in the overall system. We imagine this could be applied to a number of tasks where resource allocation is time critical.

Furthermore, the probability of waiting could be associated to some internal parameter in the system (e.g., battery level or prioritizing goals). Presumably, the impulsiveness signal in the rodent is closely tied to its natural foraging behavior. The animal will search for food, but the time it will search depends on the food value and on the uncertainty of the food resource. Such considerations could be beneficial for a robot navigation system or for a self-driving vehicle.

The road following algorithm adds another dimension to the present navigation system. By giving the robot an alternative to point-to-point navigation, the robot now must weigh the cost of staying on a smooth and reliable road that may take longer versus traversing over a rough terrain that may be shorter but takes more energy and may be harmful to the robot. Since the deep reinforcement learning introduced here is designed for online learning, these costs could be learned along with the rewards for staying on the road. Ideally, the deep reinforcement learning algorithm could set the serotonin level dynamically.

The present algorithm is a step towards a complete navigation  or self-driving system that takes inspiration from neurobiology and behavioral neuroscience.

\section{Acknowledgements}

This work was supported by the Defense Advanced Research Projects Agency (DARPA) via Air Force Research Laboratory (AFRL) Contract No. FA8750-18-C-0103 (Lifelong Learning Machines: L2M), and by the Air Force Office of Scientific Research (AFOSR) Contract No. FA9550-19-1-0306. The authors would like to thank Nicholas Ketz, Soheil Kolouri and Andrea Soltoggio for fruitful discussions and  suggestions. The authors would also like to thank Katsuhiko Miyazaki for making his code available.

\bibliographystyle{unsrt}  
\bibliography{references}  


\end{document}